\documentclass[11pt]{article}

\usepackage[dblblindworkshop, final]{neurips_2025}
\usepackage{amsmath,amssymb,amsthm}
\usepackage{nicefrac}
\usepackage{bm}
\usepackage{mathtools}
\usepackage{booktabs}
\usepackage{graphicx}
\usepackage{enumitem}
\usepackage{hyperref}
\hypersetup{colorlinks=true, linkcolor=blue, citecolor=blue, urlcolor=blue}
\usepackage{tikz}
\usetikzlibrary{arrows.meta,positioning}
\usepackage{multirow}

\newcommand{\E}{\mathbb{E}}

\newcommand{\indic}{\mathbb{I}}

\newcommand{\pr}{\mathbb{P}}

\newtheorem{definition}{Definition}
\newtheorem{assumption}{Assumption}
\newtheorem{proposition}{Proposition}

\title{Triangulation as an Acceptance Rule for\\Multilingual Mechanistic Interpretability}
\workshoptitle{Evaluating the Evolving LLM Lifecycle: Benchmarks, Emergent Abilities, and Scaling}
\author{ Yanan Long \\ StickFlux Labs \\ \texttt{ylong@uchicago.edu} }

\begin{document}
%\maketitle
\maketitle

\begin{abstract}
    Multilingual language models achieve strong aggregate performance yet often behave unpredictably across languages, scripts, and cultures. We argue that mechanistic explanations for such models should satisfy a \emph{causal} standard: claims must survive causal interventions and must \emph{cross-reference} across environments that perturb surface form while preserving meaning. We formalize \emph{reference families} as predicate-preserving variants and introduce \emph{triangulation}, an acceptance rule requiring necessity (ablating the circuit degrades the target behavior), sufficiency (patching activations transfers the behavior), and invariance (both effects remain directionally stable and of sufficient magnitude across the reference family). To supply candidate subgraphs, we adopt automatic circuit discovery and \emph{accept or reject} those candidates by triangulation. We ground triangulation in causal abstraction by casting it as an approximate transformation score over a distribution of interchange interventions, connect it to the pragmatic interpretability agenda, and present a comparative experimental protocol across multiple model families, language pairs, and tasks. Triangulation provides a falsifiable standard for mechanistic claims that filters spurious circuits passing single-environment tests but failing cross-lingual invariance.
\end{abstract}

\section{Introduction}
The success of multilingual language models has disguised a persistent pattern: large average gains mask instability across languages, writing systems, and cultures. When one isolates language- or culture-specific subsets, model rankings can invert; when inputs mix languages within a sentence, models often leak information or rely on brittle shortcuts~\citep{pfeiffer_lifting_2022}. These observations suggest that many analyses of internal states are, at best, associational: they reveal where information is encoded but not whether it \emph{causes} behavior.

We take a simple position. A mechanistic explanation should be accepted only if it remains valid under interventions and \emph{cross-references} across environments that keep meaning fixed while perturbing nuisance attributes such as language and script. Multilinguality offers exactly these environments. The literature on invariant causal prediction establishes how stability across environments can reveal causal parents~\citep{peters_causal_2016}, while causal abstraction clarifies when low-level interventions should commute with high-level semantic changes~\citep{geiger_causal_2025}. What has been missing is a standard that integrates these ingredients into an \emph{acceptance rule} for mechanism claims---one that provides empirical feedback and proxy tasks in the spirit of pragmatic interpretability~\citep{nanda_pragmatic_2025}.

We propose such a standard and call it \textbf{triangulation}. Triangulation evaluates a proposed circuit in three complementary ways. First, it requires \emph{necessity}: ablating the circuit must degrade the target behavior. Second, it requires \emph{sufficiency}: patching activations from predicate-aligned sources must transfer the behavior. Third, it requires \emph{invariance}: the sign, magnitude, and stability of causal effects must hold across a predicate-preserving reference family. In practice, this rule filters out mechanisms that owe their apparent success to language identity, script, register, or other surface cues. Triangulation is an \emph{acceptance} rule rather than a discovery method: we first propose candidate circuits via automatic discovery, then accept or reject them by triangulation.

This paper makes four contributions. We develop a formal acceptance rule grounded in causal abstraction, with testable necessity, sufficiency, and invariance criteria. We provide a protocol for constructing and auditing predicate-preserving reference families, including quality metrics and stress tests. We design a comparative experimental suite spanning multiple model families, language pairs, task types, and baseline methods, with explicit compute accounting. Finally, we extend the framework to multimodal settings where the same methodology applies with minimal adaptation.

For space, we defer related work (including the proxy-task framing) to Appendix~\ref{app:related} and the full formal framework to Appendix~\ref{app:framework}.

\section{Setup}
\label{sec:setup}
\begin{figure}[t]
	    \centering
	    \begin{tikzpicture}[ node distance=1.4cm and 1.6cm, every node/.style={draw, rounded corners, align=center, inner sep=3pt, font= \small}, arrow/.style={-{Latex[length=2mm]}, thick} ]
	        \node (E) {$E$};
	        \node[below=0.8cm of E] (Z) {$Z$};
	        \node[right=of E] (C) {$C$};
	        \node[right=of C] (X) {$X$};
	        \node[below=0.8cm of X] (V1) {$V_1$};
        \node[right=of V1] (V2) {$V_2$};
        \node[right=of V2] (Vdots) {$\cdots$};
        \node[right=of Vdots] (Vm) {$V_m$};
        \node[above=0.8cm of Vm] (Y) {$Y$};
        \node[right=of Y] (M) {$M$};

        \draw[arrow] (E) -- (C);
        \draw[arrow] (C) -- (X);
        \draw[arrow] (Z) -- (X);
        \draw[arrow] (X) -- (V1);
        \draw[arrow] (V1) -- (V2);
        \draw[arrow] (V2) -- (Vdots);
        \draw[arrow] (Vdots) -- (Vm);
        \draw[arrow] (Vm) -- (Y);
        \draw[arrow] (X) to[bend left=20] (Y);
        \draw[arrow] (Y) -- (M);
    \end{tikzpicture}
    \caption{Causal DAG for multilingual mechanisms. The experimenter selects an environment $E$ (language, script, style), which induces nuisances $C$ and shapes the realized input $X$ together with the predicate variable $Z$. The input $X$ propagates through internal states $V_1,\ldots,V_m$ to output $Y$ and task score $M$.}
    \label{fig:dag}
\end{figure}
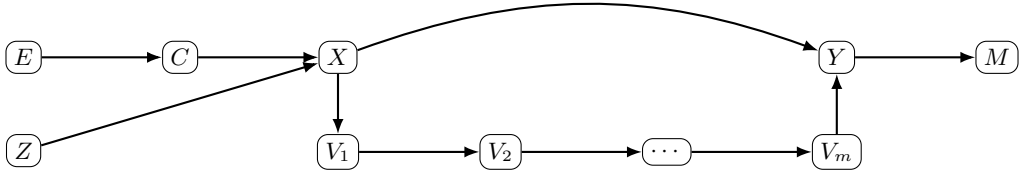

Triangulation treats multilinguality as controlled \emph{environment} variation. Fix a predicate function $\pi$ that extracts the semantic property $Z=\pi(X)$ relevant to the behavior under study (e.g., content preservation plus referent-gender constraints), and construct a \emph{reference family} $\mathcal{R}(x)=\{r_1,\ldots,r_K\}$ of predicate-preserving variants with $\pi(r_k)=\pi(x)$. We treat the index $e\in\{1,\ldots,K\}$ as an environment label capturing surface attributes (language, script, style), and test whether a proposed circuit $\mathcal{C}$ mediates behavior under interventions \emph{across} these environments. We write $a_\mathcal{C}(x)$ for the activation vector at the sites in $\mathcal{C}$ when running the model on input $x$. Figure~\ref{fig:dag} sketches the causal picture; the full SCM formalization and reference-family auditing details are in Appendix~\ref{app:framework}.

\section{Triangulation: The Acceptance Rule}
\label{sec:triangulation}
Triangulation is an \emph{acceptance rule} for mechanistic claims: given a proposed low-level mechanism, we accept it only if it behaves like an abstract predicate-level mechanism under interventions, and continues to do so across predicate-preserving environments. This section makes that claim precise by casting triangulation as an \emph{approximate causal abstraction score} computed over a distribution of \emph{interchange interventions}~\citep{geiger_causal_2025}.

\subsection{Triangulation as an Approximate Transformation Score}
\label{sec:triangulation-score}
Let $L$ denote the low-level causal model corresponding to the transformer computation and let $H$ be an abstract causal model whose key endogenous variable is the predicate $Z$ and whose behavioral output is summarized by a scalar score. For our purposes, it suffices to take $H$ to implement a simple invariant prediction: conditional on $Z$, the behavior is independent of the environment $E$.

\citeauthor{geiger_causal_2025} define \emph{approximate transformation} (Def.~41) as an expected similarity between low-level and high-level outcomes under a distribution of interventions. We instantiate this by choosing (i) a family of interventionals $\Psi$ consisting of (distributed) interchange interventions (activation patching, including null-source ablations and cue-only negative controls), (ii) a distribution $P$ over $\Psi$, (iii) an outcome-extraction map $\tau$ that returns the scalar score from a solved model, and (iv) a similarity function $\mathrm{Sim}$. The resulting triangulation score is
\begin{align}
    T_{\mathrm{tri}} := \E_{I  \sim P}\!\left[\mathrm{Sim}\!\Big(\tau\big(\mathrm{Solve}(L^I)\big),\,\tau\big(\mathrm{Solve}(H^{\omega(I)})\big)\Big)\right],
    \label{eq:tri-score}
\end{align}
where $\omega$ maps each low-level intervention $I$ to the corresponding abstract intervention on $H$ (e.g., setting $Z$ to a source predicate value). When $\mathrm{Sim}$ is an indicator of agreement on a binary outcome, Eq.~\eqref{eq:tri-score} reduces to interchange-intervention accuracy~\citep{geiger_causal_2025}.

\subsection{Interchange Interventions and Translated Patching}
\label{sec:interchange}
Activation patching is most naturally viewed as an \emph{interchange intervention}~\citep[\S 2.5]{geiger_causal_2025}. Given a base input $x_b$ (environment $e_b$) and a source input $x_s$ (environment $e_s$), we intervene on a nominated set of internal sites (a circuit) and replace their values in the base run with the values they would take in the source run. When $e_s \neq e_b$ (e.g., cross-language patching), naive value replacement may be off-manifold or even type-incompatible (tokenization/position mismatch). We therefore allow \emph{translated patching} via a learned map $T_{e_b \leftarrow e_s}$, corresponding to Geiger et al.'s distributed interchange interventions (Def.~46): we map source activations into the base circuit's coordinate system prior to replacement. We write $\mathcal{I}_{\mathrm{patch}}(\mathcal{C}_{e_b}; x_b \leftarrow x_s, T_{e_b \leftarrow e_s})$ for this intervention. Knockout-style ablations $\mathcal{I}_{\mathrm{KO}}(\mathcal{C}_{e_b})$ (mean ablation, resampling, or zeroing) can be regarded as a special case of distributed interchange intervention in which the source activations are drawn from a null distribution.

\subsection{Operational Criteria as a Single Score}
\label{sec:triangulation-criteria}
To make Eq.~\eqref{eq:tri-score} an \emph{acceptance rule}, we choose $P$ and $\mathrm{Sim}$ so that high score requires three behaviors: necessity, sufficiency, and invariance \emph{plus} explicit falsifiers for surface-cue mechanisms. Concretely, we take $P$ to be a mixture over intervention families and $\mathrm{Sim}$ to be an indicator that the observed score change lands in a preregistered acceptance region.

\paragraph{Thresholds.}
We write $\tau_N$ for the minimum necessity drop under knockout, $\tau_S$ for the minimum causal transfer under predicate-swap patching, $\epsilon$ for tolerable changes under invariant/stability tests and cue-only negative controls, and $\delta$ for an on-manifold bound (maximum allowed activation distortion under translated patching). We write $\eta \in (0,1)$ for the minimum triangulation score required for acceptance and $\alpha \in (0,1)$ for a target placebo acceptance rate used in calibration (\S\ref{sec:estimation}).

\paragraph{Necessity interventions.}
Sample $(x,e) \sim \mathcal{D}$ and apply $\mathcal{I}_{\mathrm{KO}}(\mathcal{C}_e)$. Define
\begin{align}
    \Delta_{\mathrm{KO}}(x,e) := M(f(x)) - M\bigl(f^{\mathcal{I}_{\mathrm{KO}}(\mathcal{C}_e)}(x)\bigr).
    \label{eq:necessity}
\end{align}
The necessity similarity is $\mathrm{Sim}_N = \indic[\Delta_{\mathrm{KO}}(x,e)\ge \tau_N]$.

\paragraph{Sufficiency (predicate-swap) interventions.}
Sample a base $(x_b,e_b)$ and a source $(x_s,e_s)$ with \emph{different} predicate values $\pi(x_s)\neq \pi(x_b)$, then patch the nominated circuit in the base environment:
\begin{align}
    \Delta_{\mathrm{swap}}(x_b,x_s;e_b,e_s) := M\!\left(f^{\mathcal{I}_{\mathrm{patch}}(\mathcal{C}_{e_b}; x_b \leftarrow x_s, T_{e_b\leftarrow e_s})}(x_b)\right) - M(f(x_b)).
    \label{eq:sufficiency}
\end{align}
Let $s(\pi(x_s),\pi(x_b)) \in \{-1,+1\}$ denote the predicted direction of change under the abstract model (e.g., inclusive $\to$ larger score than binary). The sufficiency similarity is $\mathrm{Sim}_S=\indic[s(\pi(x_s),\pi(x_b))\cdot \Delta_{\mathrm{swap}}\ge \tau_S \;\wedge\; \|T_{e_b\leftarrow e_s}a_{\mathcal{C}_{e_s}}(x_s)-a_{\mathcal{C}_{e_b}}(x_b)\|\le \delta]$.

\paragraph{Invariance and surface-cue falsification.}
Triangulation is meant to reject mechanisms whose apparent causal effect is carried by nuisance cues (language identity, script, punctuation, register) rather than the predicate. We therefore include two families of interventions:
\begin{enumerate}[label=(\roman*)]
    \item \emph{Stability}: patch between \emph{predicate-matched} sources, $\pi(x_s)=\pi(x_b)$, across environments, and require $|\Delta_{\mathrm{swap}}(x_b,x_s;e_b,e_s)|\le \epsilon$.
    \item \emph{Cue-only falsifiers}: patch nominated cue circuits (e.g., language-ID features) while holding the predicate circuit fixed, and require the induced score change to be small, $|\Delta_{\mathrm{cue}}|\le \epsilon$.
\end{enumerate}
Formally, for a cue circuit $\mathcal{C}^{\mathrm{cue}}_{e_b}$ and source $(x_s,e_s)$, define
\begin{align}
    \Delta_{\mathrm{cue}}(x_b,x_s;e_b,e_s) := M\!\left(f^{\mathcal{I}_{\mathrm{patch}}(\mathcal{C}^{\mathrm{cue}}_{e_b}; x_b \leftarrow x_s, T_{e_b\leftarrow e_s})}(x_b)\right) - M(f(x_b)).
    \label{eq:cue-only}
\end{align}
Both are encoded via an indicator similarity $\mathrm{Sim}_I$.

\paragraph{Score and decision rule.}
Let $\mathrm{Sim}$ be the conjunction of the relevant indicator similarities for the sampled intervention (necessity, swap, stability, or cue-only). Then $T_{\mathrm{tri}}$ in Eq.~\eqref{eq:tri-score} is the probability (under $P$) that the mechanism behaves as predicted by the abstract model. We accept a proposed mechanism class if $T_{\mathrm{tri}} \ge \eta$ \emph{and} the score remains high when conditioning on each base environment (mechanism-class invariance; \S\ref{sec:mech-class}).

\subsection{Mechanism-Class Invariance Across Environments}
\label{sec:mech-class}
A key lesson from recent multilingual circuit analyses is that the \emph{same abstract computation} can be implemented by a mixture of shared and language-specific components rather than a single fixed subgraph~\citep{lindsey_biology_2025}.

\begin{definition}[Mechanism class]
    A \emph{mechanism class} is an environment-indexed family of candidate circuits $\{\mathcal{C}_e\}_{e\in\mathcal{E}}$, optionally augmented with translation maps $\{T_{e\leftarrow e'}\}$ used for translated patching across environments.
\end{definition}

Accordingly, we treat the object under test as a mechanism class $\mathfrak{C}$ rather than a single environment-specific circuit. Triangulation tests whether $\mathfrak{C}$ supports a stable abstraction-level intervention semantics, even if the low-level implementation differs across environments.

Operationally, we compute environment-conditional scores
\begin{align}
    T_{\mathrm{tri}}(e) := \E_{I \sim P(\cdot \mid e_b=e)}[\mathrm{Sim}(\cdot)]
    \label{eq:tri-by-env}
\end{align}
and require $\min_{e\in\mathcal{E}} T_{\mathrm{tri}}(e) \ge \eta$. This criterion allows language-specific ``adapter'' subcircuits as long as (i) predicate-swap effects transfer with the correct direction/magnitude after translated patching, and (ii) cue-only falsifiers do \emph{not} transfer the behavior.

\subsection{Estimation and Calibration}
\label{sec:estimation}
Given sampled interventions $I_1,\ldots,I_n  \sim P$, define Bernoulli variables
\begin{align}
    S_i := \mathrm{Sim}(I_i)\in\{0,1\}, \qquad \hat{T}_{\mathrm{tri}} := \frac{1}{n} \sum_{i=1}^n S_i.
    \label{eq:tri-estimator}
\end{align}
Under i.i.d.\ sampling, $T_{\mathrm{tri}}=\E[S_i]$ and $k:= \sum_{i=1}^n S_i  \sim \mathrm{Binomial}(n,T_{\mathrm{tri}})$, with $\hat{T}_{\mathrm{tri}}=k/n$ the empirical estimate.

To quantify uncertainty in $T_{\mathrm{tri}}$, we use a conjugate Beta--Binomial model: place a prior $T_{\mathrm{tri}} \sim \mathrm{Beta}(a_0,b_0)$ and update with $k$ successes to obtain
\begin{align}
    T_{\mathrm{tri}} \mid \{S_i\}_{i=1}^n  \sim \mathrm{Beta}(a_0+k,\,b_0+n-k),
    \label{eq:beta-posterior}
\end{align}
from which we report posterior means and equal-tailed credible intervals given by Beta quantiles (e.g., with $a_0=b_0=1$ or Jeffreys prior $a_0=b_0=\nicefrac{1}{2}$). Appendix~\ref{app:estimation} provides full mathematical details and notation. To set $\eta$ for a target placebo acceptance rate $\alpha$, we calibrate using placebo circuits (random circuits or circuits optimized for nuisance prediction) and choose $\eta$ such that the acceptance rate on placebos is approximately $\alpha$ under the full intervention distribution $P$.

\section{Experimental Protocol}
\label{sec:experiments}
\subsection{Models}
We propose to evaluate triangulation on multiple model families that are compatible with activation patching and have demonstrated multilingual capabilities. The first family is Gemma~3\footnote{\url{https://huggingface.co/google/gemma-3-4b-it}} (1B and 4B variants) which are supported via Penzai's Gemma variant tooling and offer a good balance between capability and computational tractability. The second family is Llama~3.2\footnote{\url{https://huggingface.co/meta-llama/Llama-3.2-3B-Instruct}} in its 1B and 3B parameter variants, which are explicitly positioned as multilingual models and can be converted to Penzai format using available utilities. The third family is Mistral 3\footnote{\url{https://huggingface.co/mistralai/Ministral-3-3B-Base-2512}} in its 3B variant, which is supported via Penzai's Mistral conversion and provides an additional architectural comparison point. Finally, we include MADLAD-400-3B-MT\footnote{\url{https://huggingface.co/google/madlad400-3b-mt}}, a dedicated multilingual machine translation (MT) model based on the T5 architecture covering over 450 languages and represents ``real MT'' rather than prompted LLM translation.

\subsection{Datasets and Reference Families}
We use three translation/rewrite datasets as sources of (i) predicate instances and (ii) controlled predicate \emph{swaps} (e.g., gender requirement variants), then construct \emph{environment} reference families by varying language/script/style \emph{while holding the full predicate fixed} (Assumption~\ref{ass:predicate}).

The first is FairTranslate for English-to-French translation, which contains 2{,}418 sentence pairs where each underlying proposition appears in multiple gender-marked variants~\citep{jourdan2025fairtranslate}. Crucially, we treat these gender variants as \emph{different predicate values} for predicate-swap interventions (triangulation sufficiency), not as reference-family members. To obtain predicate-preserving environments, we translate each variant into additional source languages (e.g., German/Spanish/Italian) via human translation or high-quality MT and audit predicate preservation with checkers and annotators.

The second is GLITTER for English-to-German translation, a multi-reference benchmark with post-edited translations supporting multiple gender-fair strategies (e.g., gender-star notation vs.\ neutralization)~\citep{pranav2025glitter}. Here we treat the strategy choice as an environment axis: the predicate (content + referent constraints) is fixed while surface realization varies systematically.

The third is mGeNTE, built on Europarl, spanning English to Italian, Spanish, German, and Greek and providing both gender-ambiguous and gender-unambiguous items plus intra-lingual rewriting support~\citep{savoldi2025mind}. We use its parallel structure to construct cross-lingual reference families directly (same predicate, different source language), and its gender conditions to define predicate-swap sources.

For each dataset, we define the predicate $\pi$ to include content preservation \emph{and} any semantic requirements relevant to the target behavior (e.g., referent gender constraints in gendered translation). We implement predicate checkers combining rule-based and human/LLM-judge approaches, report inter-checker agreement as a quality metric, and run stress tests by injecting predicate violations (entity swaps, meaning flips) to verify triangulation rejects when Assumption~\ref{ass:predicate} fails.

\subsection{Tasks}
We consider two task types that exercise different aspects of multilingual processing while cleanly separating predicate variation from environment variation.

\paragraph{Translation with fixed target language.}
We fix a target language (e.g., French) and vary the source language/environment within a reference family $\mathcal{R}(z)$ that preserves predicate $z=\pi(x)$. The behavioral score $M$ is defined as a logit margin for inclusive versus binary gender realization at the target-side gender-bearing locus (GBL): the earliest decoding position where the target language commits to gender marking. Predicate-swap sources (e.g., masculine vs.\ inclusive requirements for the same underlying proposition) define the ``swap'' interventions used to test sufficiency. See App.~\ref{sec:case-study} for a worked example.

\paragraph{Intra-lingual rewriting.}
Using mGeNTE, we study rewriting within a language while varying surface form (style/register) as the environment. The behavioral score $M$ is a neutrality classifier (or logit margin) measuring whether the rewritten output avoids unnecessary gender marking while preserving the predicate.

\subsection{Baselines}
We compare triangulation acceptance against four baseline methods to assess its value-added over existing approaches. The first baseline is single-environment patching, which applies standard activation patching without cross-reference checks and represents the most common current practice. The second baseline is causal mediation analysis, which decomposes effects into indirect and direct components through nominated mediators~\citep{vig_investigating_2020}. The third baseline is causal scrubbing, which uses behavior-preserving resampling to falsify mechanistic hypotheses~\citep{chan_causal_2022}. The fourth baseline is ablation-based circuit testing using faithfulness scores from ACDC-style methods~\citep{conmy_towards_2023}.

The comparison focuses on acceptance rates: methods that accept more circuits are potentially admitting more false positives (spurious circuits), while methods that accept fewer circuits may be overly conservative. The key question is whether triangulation achieves a better trade-off, rejecting circuits that fail cross-lingual invariance while retaining circuits that pass it.

\subsection{Discovery Pipeline}
For each predicate instance $z$ and its reference family $\mathcal{R}(z)$, the pipeline proceeds in three stages:
\begin{enumerate}[label=(\roman*)]
    \item \emph{Discovery.} For each environment $e\in\mathcal{R}(z)$, run an automatic discovery method (e.g., EAP-IG or position-aware circuit discovery) to propose an environment-specific circuit $\mathcal{C}_e$ for the score $M$. In parallel, learn candidate \emph{cue circuits} (language/script predictors) to serve as explicit negative controls.
    \item \emph{Translation maps.} Using predicate-matched pairs within reference families, fit translation maps $T_{e_b\leftarrow e_s}$ that align activations for translated patching (\S\ref{sec:interchange}), with strict hold-out to prevent leakage into acceptance evaluation.
    \item \emph{Acceptance.} Evaluate the resulting mechanism class $\mathfrak{C}=\{\mathcal{C}_e\}$ under the triangulation score $T_{\mathrm{tri}}$ (Eq.~\eqref{eq:tri-score}) using an intervention distribution $P$ that includes knockout, predicate-swap patching, stability tests, and cue-only falsifiers (\S\ref{sec:triangulation-criteria}). Accept only if the score is high overall and across base environments (Eq.~\eqref{eq:tri-by-env}).
\end{enumerate}

This pipeline separates discovery from acceptance, ensuring that no manual head-picking influences the results. The same discovered circuits are evaluated under all methods, enabling fair comparison of acceptance criteria.

\section{Extension to Multimodal Settings}
\label{sec:multimodal}
The triangulation framework extends to vision-language models (VLMs) by constructing reference families over the text modality while holding the image fixed.

\subsection{Multimodal SCM}
For a VLM, we take the predicate $Z$ to be an image-derived attribute (e.g., object identity) and the environment $E$ to be the prompt language/phrasing. We hold the image fixed within a reference family and vary the text realization to create multiple environments:
\begin{align}
    X &= (I, T), \quad Z = \pi(I), \quad T = g_T(E).
    \label{eq:vlm-scm}
\end{align}
The internal activations depend on both modalities:
\begin{align}
    V_i = f_i\bigl(\mathrm{Pa}(V_i), I, T\bigr).
    \label{eq:vlm-activations}
\end{align}
Here, the nuisance attributes (language, phrasing, grammatical gender markers) reside in $T$ and are controlled by $E$.

\subsection{Reference Family Construction}
For anchor input $x = (I, T_{\mathrm{en}})$ with predicate $\pi(I)$, the reference family is
\begin{align}
    \mathcal{R}(x) = \{(I, T_{\mathrm{en}}), (I, T_{\mathrm{fr}}), (I, T_{\mathrm{de}}), (I, T_{\mathrm{es}})\},
    \label{eq:vlm-ref-fam}
\end{align}
where $T_\ell$ is the prompt in language $\ell$. By construction,
\begin{align}
    \pi(I, T_\ell) = \pi(I) \quad \forall \ell,
    \label{eq:vlm-predicate-inv}
\end{align}
since the predicate depends only on the image.

\subsection{Task Score for Visual Attributes}
Let $\mathcal{A} = \{a_1, \ldots, a_L\}$ be the set of possible attribute values (e.g., professions, object categories). The task score is the log-probability of the correct attribute:
\begin{align}
    M(I, T) := \log p_\theta\bigl(\pi(I) \mid I, T\bigr) = \log  \sum_{v \in \mathcal{V}_{\pi(I)}} p_\theta(v \mid I, T),
    \label{eq:vlm-task-score}
\end{align}
where $\mathcal{V}_a  \subset \mathcal{V}$ is the set of tokens corresponding to attribute $a$.

\subsection{Triangulation Criteria for VLMs}
We apply the same triangulation score $T_{\mathrm{tri}}$ (Eq.~\eqref{eq:tri-score}) with predicate $Z=\pi(I)$ and environments given by prompt language/phrasing. In particular, we evaluate a mechanism class $\mathfrak{C}$ under an intervention distribution that includes:
\begin{enumerate}[label=(\roman*)]
    \item \emph{Necessity:} KO on the nominated visual circuit should reduce $M(I,T_\ell)$ for each prompt language $\ell$.
    \item \emph{Stability across prompt languages:} predicate-matched patching between $(I,T_\ell)$ and $(I,T_{\ell'})$ (with translated patching if needed) should have small effect on $M$.
    \item \emph{Predicate swaps:} swapping the image predicate (using a different image $I'$ with $\pi(I')\neq \pi(I)$) and patching the visual circuit should shift $M$ in the direction predicted by the change in $Z$, independent of $\ell$.
    \item \emph{Cue-only falsifiers:} patching grammatical-gender or language-ID circuits in the text pathway should not reliably induce changes in the visual-attribute score.
\end{enumerate}
This mirrors the multilingual text-only setting: we accept only mechanisms that are stable under environment changes but sensitive to predicate changes, while explicitly rejecting nuisance-cue routes.

\subsection{Disentangling Visual and Linguistic Circuits}
Triangulation enables principled separation of circuits into three categories:
\begin{align}
    \mathfrak{C}_{\mathrm{vis}} &: \quad \text{high }T_{\mathrm{tri}} \text{ across prompt languages} \;\Rightarrow\; \text{visual mechanism class}, \nonumber \\ \mathfrak{C}_{\mathrm{ling}} &: \quad \text{single-env success, cue sensitivity / stability failures} \;\Rightarrow\; \text{linguistic/cue mechanism}, \nonumber \\ \mathfrak{C}_{\mathrm{spur}} &: \quad \text{fails KO or predicate swaps} \;\Rightarrow\; \text{spurious}.
    \label{eq:circuit-taxonomy}
\end{align}
This taxonomy addresses the criticism that multilinguality-motivated frameworks are not truly multimodal: by holding the image fixed and varying text, we isolate circuits that extract visual information from those that merely correlate with linguistic cues.

\section{Discussion}
\label{sec:discussion}
Triangulation raises the evidential bar for mechanistic claims by requiring necessity, sufficiency, and cross-environment invariance. In its revised form, triangulation is an approximate causal abstraction score (Eq.~\eqref{eq:tri-score}) computed over interchange interventions, with explicit cue-only falsifiers and mechanism-class invariance across environments. This higher bar filters mechanisms that would otherwise be accepted by single-environment patching, aligning with the pragmatic interpretability goal of empirical feedback on mechanistic claims~\citep{nanda_pragmatic_2025}. The framework provides a concrete instantiation of what it means for a mechanistic explanation to be ``robust'' rather than spuriously tied to surface cues.

Several limitations merit acknowledgment. First, multiple circuits may satisfy triangulation if their effects are redundant; triangulation identifies \emph{sufficient} mechanisms but does not guarantee uniqueness. Second, the quality of reference families is decisive, and poor-quality families can mislead by either accepting spurious circuits (if predicate violations go undetected) or rejecting genuine circuits (if the family members are too dissimilar in ways that affect mechanism function). We recommend auditing reference families with multiple checkers and conducting stress tests before drawing conclusions. Third, causal interventions at scale are computationally expensive, and prioritization strategies are advisable for practical deployment. Fourth, when using LLM judges to verify predicate preservation, circularity risks arise if the judge correlates with the model under study; rule-based or human checks are preferable where feasible.

Beyond multilingual translation, triangulation applies to any setting with predicate-preserving environments. Examples include cross-lingual question answering (same question in multiple languages should yield the same answer), morphosyntactic agreement (number or tense marking should be consistent across languages with the same underlying structure), and entity preservation (names and quantities should transfer correctly regardless of surrounding linguistic context). The framework's generality stems from its grounding in causal abstraction and invariance principles rather than task-specific heuristics.

\section{Conclusion}
We have introduced triangulation, a three-part acceptance rule for mechanistic interpretability claims that requires necessity, sufficiency, and invariance across predicate-preserving reference families. Grounded in causal abstraction theory~\citep{geiger_causal_2025} and aligned with pragmatic interpretability~\citep{nanda_pragmatic_2025}, triangulation provides an objective, falsifiable criterion---implemented as an approximate transformation score under interchange interventions---for evaluating whether a proposed mechanism genuinely mediates a target behavior or merely correlates with it in a particular evaluation environment. The framework applies across multiple model families, language pairs, and task types, and extends naturally to multimodal settings.

By separating discovery from acceptance and requiring cross-environment consistency, triangulation offers a concrete step toward rigorous, transferable mechanistic explanations. Circuits that pass triangulation can be interpreted with greater confidence as capturing language-agnostic mechanisms rather than surface-dependent shortcuts. This confidence, while not absolute, represents meaningful progress toward the pragmatic interpretability goal of achieving impact through focused, falsifiable claims about model internals.

\clearpage
\bibliographystyle{unsrtnat}
\bibliography{references_MI}
\clearpage

\appendix

\section{Related Work}
\label{app:related}
\paragraph{Mechanistic interpretability and its limits.}
Interventional tools such as causal tracing and path patching underpin contemporary mechanistic interpretability for large language models~\citep{meng_locating_2022,vig_investigating_2020}, but outcomes are highly sensitive to corruption choices and evaluation metrics. This sensitivity has motivated on-manifold constraints and stricter experimental protocols~\citep{wang_interpretability_2023}. Recent evidence further demonstrates that circuit faithfulness scores can be brittle to seemingly minor ablation details, reinforcing the need for acceptance criteria that go beyond single-environment patch scores~\citep{miller_transformer_2024}. Hanna et al.\ argue persuasively that circuit overlap alone is insufficient for validating mechanistic claims~\citep{hanna_have_2024}.

\paragraph{Automatic circuit discovery.}
We use the term ``automatic circuit discovery'' broadly for pipelines that algorithmically produce candidate subgraphs with minimal manual curation. Search-based methods such as ACDC directly return sparse circuits that preserve behavior on held-out inputs~\citep{conmy_towards_2023}, while position-aware variants add token-span sensitivity and automated schema construction, improving the size--faithfulness trade-off~\citep{haklay_position-aware_2025}. Edge-scoring methods like edge attribution patching (EAP) and EAP with integrated gradients (EAP-IG) automatically rank edges by importance; coupled with selection rules such as thresholding or pruning, they also yield circuits~\citep{hanna_have_2024}. In our pipeline, automatic discovery \emph{proposes} subgraphs, and triangulation \emph{determines acceptance}.

\paragraph{Causal mediation and falsification.}
Causal mediation analysis decomposes total effects into natural indirect and direct components through nominated mediators~\citep{vig_investigating_2020,pearl_causality_2009}. Causal scrubbing offers behavior-preserving resampling tests to falsify mechanistic hypotheses~\citep{chan_causal_2022}. Our approach shares the falsification ethos but avoids cross-world assumptions by requiring \emph{invariance} of the predictive link and \emph{directional stability} of interventional effects across predicate-preserving references.

\paragraph{Invariance and causal abstraction.}
Invariant Causal Prediction formalizes why causal parents support stable conditional behavior across environments~\citep{peters_causal_2016}. Causal abstraction provides a principled account of when low-level interventions should commute with high-level changes, yielding graded faithfulness between circuits and interpretable models~\citep{geiger_causal_2025}. However, invariance alone does not identify latent causal variables~\citep{ahuja_interventional_2023}; this limitation motivates why triangulation requires \emph{both} invariance and interventional consistency.

\paragraph{Pragmatic interpretability and proxy tasks.}
The DeepMind mechanistic interpretability team has advocated a pivot toward pragmatic interpretability: grounding research with proxy tasks, pursuing theories of change, and exercising method minimalism~\citep{nanda_pragmatic_2025,nanda_how_2025}. Our triangulation rule operationalizes this philosophy by providing an objective, falsifiable acceptance criterion, and it can be framed directly as a proxy task: given a proposed mechanism class, estimate $T_{\mathrm{tri}}$ (Eq.~\eqref{eq:tri-score}) and test whether the mechanism behaves like an abstract predicate mechanism under a preregistered distribution of interchange interventions across predicate-preserving environments. This proxy returns an empirical score with explicit falsifiers (cue-only interventions) and stress tests (intentional predicate violations), providing feedback on whether mechanistic claims survive contact with reality. Triangulation supports the ``science of misalignment'' theory of change: if a model misbehaves, we can rigorously investigate whether the responsible circuit is language-specific (and therefore spurious) or language-agnostic (and therefore a genuine mechanism), helping distinguish scheming from confusion.

\section{Formal Framework}
\label{app:framework}
This appendix provides the full formal framework summarized in \S\ref{sec:setup} and Figure~\ref{fig:dag}.

\subsection{Structural Causal Model}
\label{sec:scm}
We model a model forward pass as a structural causal model (SCM) with two \emph{exogenous} variables: a predicate variable $Z$ (the semantic property we wish to track) and an environment index $E$ (language, script, register, formatting, etc.). The environment $E$ is selected by the experimenter via choice of a reference-family member; it is not an endogenous ``cause'' of the data in the usual sense, but it plays the same mathematical role as the environment variable in invariant causal prediction. To keep contact with common mechanistic interpretability practice, we also introduce a nuisance summary $C$ for surface attributes induced by $E$.

The endogenous variables are the realized input $X$, a collection of internal activations $V = (V_1,\ldots,V_m)$ at specified sites (heads, MLP channels, residual-stream subspaces, or feature bases), the output $Y$, and a derived task score $M=M(Y)$ (e.g., logit margin or accuracy).

The structural equations take the form
\begin{align}
    C &= g_C(E), \quad X = g_X(Z, C), \quad V_i = f_i\bigl(\mathrm{Pa}(V_i), X\bigr), \quad Y = f_Y(V, X), \quad M = m(Y),
    \label{eq:scm}
\end{align}
where $\mathrm{Pa}(V_i)$ denotes causal parents of $V_i$ among preceding internal variables $(V_1,\ldots,V_{i-1})$. We omit exogenous noise terms since, conditional on $X$, transformer forward passes are typically deterministic.

\begin{assumption}[Predicate separation]
    \label{ass:predicate}
    There exists a predicate function $\pi$ such that $Z=\pi(X)$ and $\pi$ is invariant under changes to $E$ within a reference family. That is, for any anchor input $x$ and its reference family $\mathcal{R}(x)=\{r_1,\ldots,r_K\}$, we have $\pi(r_k)=\pi(x)$ for all $k$.
\end{assumption}

Under this assumption, a reference family varies $E$ (and thus $C$) while holding $Z$ fixed. This formalizes the intuition ``same meaning, different surface form''; importantly, the predicate $\pi$ must include \emph{all} semantic constraints relevant to the behavior under study (e.g., referent gender requirements in gendered translation), otherwise ``reference families'' will silently violate Assumption~\ref{ass:predicate}.

\begin{definition}[Candidate circuit]
    A \emph{candidate circuit} $\mathcal{C}  \subseteq \{1, \ldots, m\}$ is a selected set of internal variables and possibly edges connecting them. We write $a_\mathcal{C}(x)$ for the concatenated activation vector at the sites in $\mathcal{C}$ on input $x$.
\end{definition}

\begin{definition}[Mechanistic hypothesis]
    A \emph{mechanistic hypothesis} posits a high-level causal model $H$ over abstract variables (including the predicate $Z$ and a behavioral variable derived from $M$) and an alignment $\langle \Pi,\pi\rangle$ relating the low-level transformer SCM $L$ to $H$ such that $H$ is a (possibly approximate) causal abstraction of $L$ in the sense of causal abstraction theory~\citep{geiger_causal_2025}.
\end{definition}

In Geiger et al.'s framework, exact abstraction is expressed as a commuting condition (an \emph{exact transformation}) between low-level and high-level models under a mapping $(\tau,\omega)$ over interventionals, and graded faithfulness is quantified by an \emph{approximate transformation} score computed over a distribution of interventions~\citep{geiger_causal_2025}. Triangulation is our operational acceptance rule for this graded notion: we define a distribution over \emph{interchange interventions} (activation patching, including translated patching across environments) and accept only those candidate mechanisms whose observed interventional behavior matches the predictions of an abstract model \emph{across environments} that preserve $Z$.

\begin{proposition}[Informal]
    If a proposed mechanism passes triangulation with sufficiently strong thresholds across environments, then there exists evidence (relative to the chosen intervention distribution and similarity metric) that the associated abstract model is an $\eta$-approximate causal abstraction of the low-level transformer computation in the sense of approximate transformation~(cf.\ \citep[\S 2.4]{geiger_causal_2025}).
\end{proposition}

\subsection{Reference Families as Environments}
\label{sec:reference-families}
\begin{definition}[Reference family]
    For anchor input $x$ and predicate $\pi$, a \emph{reference family} is a set $\mathcal{R}(x) = \{x^{(1)}, \ldots, x^{(K)}\}$ such that $\pi(x^{(k)}) = \pi(x)$ for all $k \in \{1, \ldots, K\}$. We treat the environment index $e \in \{1, \ldots, K\}$ as selecting which family member is used.
\end{definition}

The quality of reference families is decisive for triangulation. We operationalize predicate preservation through three complementary mechanisms. The first is a \emph{predicate checker} $c_\pi(x, x')$ that returns $1$ if $\pi(x) = \pi(x')$ and $0$ otherwise. For structured predicates such as entity identity, rule-based checkers suffice: one extracts entities from both inputs via named entity recognition and verifies set equality. For semantic equivalence, one may use an LLM judge prompted to determine whether two inputs convey the same meaning ignoring differences in language or style, ideally with chain-of-thought reasoning. For stress tests and low-resource language pairs, expert human annotators with inter-rater reliability checks (targeting $\kappa \geq 0.8$) provide the most reliable ground truth, though at greater cost.

The second mechanism is a \emph{reference-family quality score} that aggregates agreement among checkers, including inter-annotator agreement where human annotation is involved. The third mechanism is a \emph{stress test} in which we intentionally introduce small predicate violations---such as changing named entities or altering key semantic content---and verify that triangulation rejects accordingly. This stress test validates that the acceptance rule is sensitive to genuine predicate mismatches rather than merely filtering based on surface statistics.

Building high-quality reference families is non-trivial, particularly for languages with less-resourced cultural or grammatical variation. We address this limitation by reporting measured degradation in reference-family quality as we move beyond high-resource pairs, and by analyzing how robust triangulation remains as a function of that quality. When using LLM judges to verify predicate preservation, one must be cautious about circularity: if the judge belongs to the same model family as the model under study, correlations may inflate apparent reference-family quality. We therefore prefer rule-based or human checks where possible, reserving LLM judges for cases where semantic equivalence is difficult to verify automatically.

\section{Estimation and Calibration Details}
\label{app:estimation}
This appendix expands the Bayesian uncertainty-quantification statement in \S\ref{sec:estimation}.

\paragraph{Parameter.}
Let $P$ be the preregistered distribution over low-level interventions $I\in\Psi$ (including any randomization over base/source examples, environments, and intervention types). For a sampled intervention $I \sim P$, define the binary acceptance indicator
\begin{align}
    S := \mathrm{Sim}(I)\in\{0,1\}.
    \label{eq:sim-indicator}
\end{align}
Then the triangulation score is the Bernoulli mean (equivalently, a probability):
\begin{align}
    T_{\mathrm{tri}} := \E_{I \sim P}[\mathrm{Sim}(I)] = \E[S] = \pr(S=1).
    \label{eq:ttri-prob}
\end{align}

\paragraph{Sampling model and likelihood.}
Draw i.i.d.\ interventions $I_1,\ldots,I_n \overset{\mathrm{i.i.d.}}{ \sim} P$ and set $S_i:=\mathrm{Sim}(I_i)\in\{0,1\}$. Under i.i.d.\ sampling,
\begin{align}
    S_i \mid T_{\mathrm{tri}}  \sim \mathrm{Bernoulli}(T_{\mathrm{tri}}), \qquad k :=  \sum_{i=1}^n S_i \,\Big|\, T_{\mathrm{tri}}  \sim \mathrm{Binomial}(n,T_{\mathrm{tri}}).
    \label{eq:binom-model}
\end{align}

\paragraph{Beta--Binomial posterior (conjugate Bayesian uncertainty quantification).}
Treat $T_{\mathrm{tri}}$ as an unknown parameter $\theta\in(0,1)$ and place a Beta prior
\begin{align}
    \theta := T_{\mathrm{tri}}  \sim \mathrm{Beta}(a_0,b_0), \qquad a_0,b_0>0.
    \label{eq:beta-prior}
\end{align}
Together with the Binomial likelihood in \eqref{eq:binom-model}, conjugacy yields the posterior
\begin{align}
    \theta \mid \{S_i\}_{i=1}^n  \sim \mathrm{Beta}(a_0+k,\;b_0+n-k).
    \label{eq:beta-post}
\end{align}
The posterior mean is
\begin{align}
    \E[\theta\mid \{S_i\}] = \frac{a_0+k}{a_0+b_0+n},
    \label{eq:beta-mean}
\end{align}
and an equal-tailed $(1-\alpha)$ credible interval is given by Beta quantiles:
\begin{align}
    \mathrm{CI}^{\mathrm{Bayes}}_{1-\alpha} := \Bigl[ F^{-1}_{\mathrm{Beta}(a_0+k,b_0+n-k)}(\alpha/2),\; F^{-1}_{\mathrm{Beta}(a_0+k,b_0+n-k)}(1-\alpha/2) \Bigr],
    \label{eq:beta-ci}
\end{align}
where $F^{-1}_{\mathrm{Beta}(a,b)}$ is the inverse CDF of a $\mathrm{Beta}(a,b)$ distribution. Common choices are a uniform prior $(a_0,b_0)=(1,1)$ or the Jeffreys prior $(a_0,b_0)=(\nicefrac{1}{2},\nicefrac{1}{2})$.

\paragraph{Remark on dependence.}
If the sampled indicators are not i.i.d.\ (e.g., many interventions share the same anchor predicate instance), then the i.i.d.\ Binomial model in \eqref{eq:binom-model} can understate uncertainty. In that case, one can use a hierarchical Beta--Binomial model that pools within anchors (or other independent units) and reports posterior uncertainty at the appropriate level; we treat i.i.d.\ sampling as a simplifying assumption for exposition.

\section{Case Study: Inclusive English-to-French Translation}
\label{sec:case-study}
We illustrate triangulation on inclusive gender marking in English-to-French translation.

\subsection{Reference Family Construction}
Let $z=(z_{\mathrm{sem}}, z_g)$ denote a full predicate instance consisting of semantic content $z_{\mathrm{sem}}$ and a referent-gender requirement $z_g \in \{\mathrm{m},\mathrm{f},\mathrm{incl},\mathrm{amb}\}$ that is relevant for gendered French translation. We take $\pi(x)=z$; in particular, changing a pronoun from ``he'' to ``she'' changes $z_g$ and is therefore a \emph{predicate change}, not an environment change.

We construct a predicate-preserving reference family by varying the source language (and prompt style) while holding $z$ fixed:
\begin{align}
    \mathcal{R}(z) = \{r_{\mathrm{en}}, r_{\mathrm{de}}, r_{\mathrm{es}}\} \quad\text{s.t.}\quad \pi(r_{\mathrm{en}})=\pi(r_{\mathrm{de}})=\pi(r_{\mathrm{es}})=z.
    \label{eq:ref-fam}
\end{align}
Separately, for each environment $\ell\in\{\mathrm{en},\mathrm{de},\mathrm{es}\}$ we construct predicate-swap siblings $r_{\ell}^{(g)}$ that keep $z_{\mathrm{sem}}$ fixed but change $z_g$ (e.g., by minimal edits to gender cues). These swaps provide the sources for the predicate-swap interventions used to test sufficiency.

\subsection{Task Score}
Let $t \in \{1,\ldots, L\}$ index the gender-bearing locus (GBL), the earliest decoding position where French commits to gender for a length-$L$ target. Let $\mathcal{V}$ denote the model vocabulary, and define the inclusive, masculine, and feminine token sets $\mathcal{T}_{\mathrm{incl}}, \mathcal{T}_{\mathrm{m}}, \mathcal{T}_{\mathrm{f}}  \subset \mathcal{V}$. The task score is the log-odds favoring inclusive realization:
\begin{align}
    M(x) := \log  \sum_{v \in \mathcal{T}_{\mathrm{incl}}} p_\theta(v \mid x, y_{<t}) - \log \max\!\left\{  \sum_{v \in \mathcal{T}_{\mathrm{m}}} p_\theta(v \mid x, y_{<t}),  \sum_{v \in \mathcal{T}_{\mathrm{f}}} p_\theta(v \mid x, y_{<t}) \right\},
    \label{eq:task-score-gender}
\end{align}
where $p_\theta(\cdot\mid x,y_{<t})$ denotes the model's next-token distribution when translating input $x$ into French under a fixed decoding prefix $y_{<t}$ (teacher-forced or model-generated). Positive $M$ indicates preference for inclusive forms (e.g., ``iel''), negative $M$ for binary forms.

\subsection{Discovery via Edge Attribution}
For each environment $\ell$, let $S_\ell$ denote the source-side token span that realizes the gender cue(s) relevant to $z_g$ in that language. Let $\mathcal{G}_\ell$ denote the set of edges in the computational graph from $S_\ell$ to the GBL. For each edge $e \in \mathcal{G}_\ell$, we compute the integrated gradient attribution:
\begin{align}
    \mathrm{IG}(e) := (a_e - \bar{a}_e) \cdot \int_0^1 \frac{\partial M}{\partial a_e}\bigg|_{a_e = \bar{a}_e + \alpha(a_e - \bar{a}_e)} d\alpha,
    \label{eq:eap-ig}
\end{align}
where $a_e$ is the activation along edge $e$ and $\bar{a}_e$ is a baseline (e.g., mean activation). We rank edges by $|\mathrm{IG}(e)|$ and greedily prune to obtain an environment-specific circuit $\mathcal{C}_\ell$ such that
\begin{align}
    \frac{\bigl|M(f^{\mathcal{C}_\ell}(x)) - M(f(x))\bigr|}{\bigl|M(f(x))\bigr| + \varepsilon_M} \leq 0.1,
    \label{eq:faithfulness}
\end{align}
where $f^{\mathcal{C}_\ell}$ denotes the model restricted to circuit $\mathcal{C}_\ell$. Here $\varepsilon_M>0$ is a small constant (e.g., $10^{-6}$) that avoids division by zero when $M(f(x))\approx 0$. Across environments, we fit translation maps $T_{\ell\leftarrow \ell'}$ on predicate-matched pairs $(r_\ell,r_{\ell'})$ to enable translated patching.

\subsection{Triangulation Application}
We evaluate the mechanism class $\mathfrak{C}=\{\mathcal{C}_\ell\}$ using the triangulation score $T_{\mathrm{tri}}$ (Eq.~\eqref{eq:tri-score}) under an intervention distribution $P$ that mixes:
\begin{enumerate}[label=(\roman*)]
    \item \emph{Necessity (KO):} apply $\mathcal{I}_{\mathrm{KO}}(\mathcal{C}_\ell)$ on $(r_\ell,z)$ and require $\Delta_{\mathrm{KO}}(r_\ell,\ell)\ge \tau_N$ (Eq.~\eqref{eq:necessity}).
    \item \emph{Predicate swap:} patch from a predicate-swap source $r_{\ell'}^{(g')}$ into a base $r_{\ell}^{(g)}$ using translated patching $T_{\ell\leftarrow \ell'}$ and require the score shift to match the predicted direction/magnitude (Eq.~\eqref{eq:sufficiency}).
    \item \emph{Stability across environments:} patch between predicate-matched members of $\mathcal{R}(z)$ and require $|\Delta_{\mathrm{swap}}|\le \epsilon$.
    \item \emph{Cue-only falsifiers:} patch language-ID / script circuits while holding the predicate circuit fixed and require $|\Delta_{\mathrm{cue}}|\le \epsilon$.
\end{enumerate}
We accept only if $T_{\mathrm{tri}}$ is high overall and when conditioning on each base environment (Eq.~\eqref{eq:tri-by-env}).

\subsection{Expected Outcomes}
Genuine predicate mechanisms should (i) show large KO drops, (ii) transfer under predicate swaps with the correct direction/magnitude even when the source language differs (after translation maps), and (iii) remain stable under predicate-matched cross-environment patching. In contrast, nuisance-sensitive routes (language-ID or formatting cues) may appear helpful in a single environment but should fail cue-only falsifiers and stability checks, reducing $T_{\mathrm{tri}}$.

We report diagnostics that disentangle failure modes:
\begin{align}
    \text{Lexicalization failure:} &\quad \mathcal{T}_{\mathrm{incl}} \cap \mathrm{supp}(p_\theta) = \emptyset, \nonumber \\ \text{Agreement failure:} &\quad M(x) > 0 \;\wedge\; g(\hat{y}) \in \{\mathrm{m}, \mathrm{f}\}, \nonumber \\ \text{Cue sensitivity:} &\quad |\Delta_{\mathrm{cue}}| > \epsilon \text{ or stability violations across } \mathcal{R}(z).
    \label{eq:error-tax}
\end{align}

\end{document}